\documentclass[11pt]{article}

\usepackage[final]{acl}

\usepackage{times}
\usepackage{latexsym}
\usepackage{amssymb}
\usepackage{amsmath}

\usepackage[T1]{fontenc}
\usepackage[utf8]{inputenc}

\usepackage{microtype}
\usepackage{inconsolata}
\usepackage{graphicx}
\usepackage{booktabs}

\usepackage{algorithm}
\usepackage{algpseudocode}

\usepackage{xcolor}
\usepackage{tikz}
\usetikzlibrary{arrows.meta,positioning,fit,backgrounds,calc}

\newcommand{\samethanks}[1][\value{footnote}]{\footnotemark[#1]}

\title{Latent Recurrent Thoughts: Recurrent Refinement of Proposed Latents\\ for Reasoning with Frozen LLMs}

\author{Zhaoliang Chen\thanks{\;Correspondence to \texttt{zhaolc.david@gmail.com} and \texttt{jie.fu@iquestlab.com}} \\
  Emory University \\\And
  Jie Fu\samethanks \\
  IQuest Research
  }

\begin{document}
\maketitle

\begin{abstract}
Chain-of-thought reasoning unfolds in discrete token space: each step is committed as text, errors propagate, and eliciting good traces presupposes traces to imitate. Reasoning instead in a model's continuous representation space -- where intermediate states are vectors rather than words -- sidesteps these constraints, but leaves open how those latent states should be computed. We approach this along two axes. First, we keep a large language model (LLM) \emph{frozen} and use it for what it is already good at -- modeling and decoding sequences -- while a small auxiliary network supplies continuous latent thoughts as input. Second, we produce those latents by \emph{recurrence}: a tiny recurrent reasoner refines them over many steps, decoupling the depth of computation from the size of the model, so that the latents are a product of iterative processing rather than a single forward pass. We instantiate this as \textbf{Latent Recurrent Thoughts (LRT)}\footnote{Code is available at \url{https://github.com/czl-david/latent-recurrent-thoughts}.}: a task-dedicated proposer supplies base latents, a recurrent reasoner refines them through bounded residual corrections, and the frozen LLM decodes the answer. On symbolic reasoning with answer supervision but no reasoning traces (Countdown-4, Sudoku) and on natural-language reasoning (HumanEval, MBPP, StrategyQA), LRT substantially outperforms prior frozen-decoder continuous-space reasoning methods under an identical decoder, prompt, data, and training budget, and outperforms non-thinking-mode chain-of-thought prompting on the same backbone at a small fraction of its inference compute.
\end{abstract}

\section{Introduction}

Large language models (LLMs) solve complex problems by externalizing intermediate steps as chain-of-thought (CoT) traces \citep{wei2022chain,kojima2022large,nye2021show}, and recent systems push this further with sampling, search, and reinforcement learning \citep{wang2023selfconsistency,yao2023tree,besta2024graph,snell2024scaling,guo2025deepseek}. But CoT reasons in the discrete token space: each step must be verbalized from a fixed vocabulary, generation is autoregressive and error-propagating, and -- most limiting -- eliciting good CoT presupposes \emph{reasoning traces} to imitate \citep{zelikman2022star,brown2020language}. This motivates reasoning that instead operates in a model's continuous representation space.

Such \emph{continuous-space} reasoning feeds latent vectors back into the model rather than decoded tokens \citep{hao2024training,cheng2024compressed,su2025token,shen2025codi}, and existing methods realize it in many ways. One practical path -- attractive when fine-tuning a capable LLM is too costly or risks catastrophic forgetting -- keeps the decoder \emph{frozen}: a small auxiliary model produces instance-conditioned latents that are injected into its input as soft tokens -- the interface used by prefix and prompt tuning \citep{li2021prefix,lester2021power}. SoftCoT \citep{xu2025softcot} and EBM-CoT \citep{chen2025think} take this route, and so does our method.

Within this frozen-decoder approach, two components govern quality: the \emph{proposer} that supplies the latents and the \emph{refiner} that improves them. SoftCoT proposes with a frozen \emph{generic} language assistant and performs no refinement at all; EBM-CoT keeps that proposer and adds a refiner that descends a learned scalar \emph{energy} field, nudging latents toward lower-energy, ``consistent'' regions. We find both fragile. Off the natural-language substrate, a generic assistant's proposed latents can be \emph{worse than no latents} -- they actively mislead the decoder -- whereas a small proposer dedicated to the task family helps. And an energy refiner only follows the gradient of a fixed scalar field: it \emph{calibrates} latents rather than \emph{computing} with them, too shallow for multi-step search.

We introduce \textbf{Latent Recurrent Thoughts (LRT)}, which strengthens both links. A task-dedicated proposer, conditioned on the input $x$, supplies base latents adapted to the task family; a \emph{recurrent reasoner} then refines them -- we use TRM \citep{jolicoeurmartineau2025less}, a tiny recursive network that iteratively updates a latent state. Unlike an energy model, a recurrent reasoner applies a learned \emph{vector-valued} transition that can implement constraint propagation and iterative correction, and can be unrolled to greater depth at no extra parameter cost. It predicts \emph{bounded residual corrections} rather than regenerating the latent, keeping refinement anchored to the proposed base. Such recurrent reasoners -- HRM, TRM, GRAM, EqR \citep{wang2025hierarchical,jolicoeurmartineau2025less,baek2026generative,huang2026equilibrium} -- have proven to be powerful and parameter-efficient computation engines, but they have so far been built only as \emph{standalone solvers} trained from scratch on a single task. LRT instead pairs one with a frozen, pretrained LLM: the recurrent reasoner carries the iterative computation, while the LLM contributes the broad sequence-modeling and language ability that a from-scratch solver lacks. To our knowledge, this pairing is new.

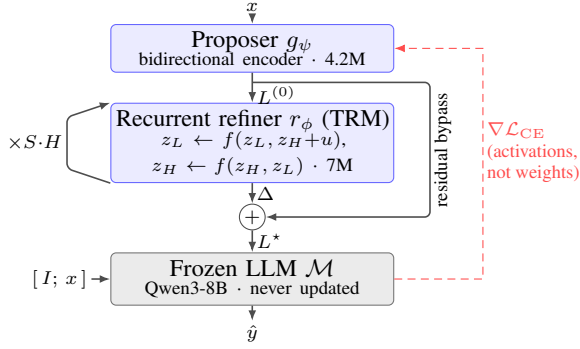
\begin{figure}[t]
\centering
\begin{tikzpicture}[
  x=1cm, y=1cm,
  every node/.style={font=\footnotesize, inner sep=2pt},
  tbox/.style={draw=blue!60, fill=blue!8, rounded corners=2pt, align=center,
               text width=36mm, inner ysep=2.5pt},
  fbox/.style={draw=black!55, fill=black!8, rounded corners=2pt, align=center,
               text width=36mm, inner ysep=2.5pt},
  lbl/.style={font=\scriptsize, inner sep=1.4pt},
  ar/.style={-{Latex[length=1.4mm]}, semithick, draw=black!70}
]
\node[lbl]  (x)    at (0, 0)     {$x$};
\node[tbox] (prop) at (0,-0.55)  {Proposer $g_\psi$\\[-3pt]{\scriptsize bidirectional encoder $\cdot$ 4.2M}};
\node[tbox] (ref)  at (0,-1.80)  {Recurrent refiner $r_\phi$ (TRM)\\[-3pt]{\scriptsize $z_L\!\leftarrow\!f(z_L,z_H{+}u)$, $z_H\!\leftarrow\!f(z_H,z_L)$ $\cdot$ 7M}};
\node[circle, draw=black!70, inner sep=0.6pt] (plus) at (0,-2.78) {\scriptsize$+$};
\node[fbox] (llm)  at (0,-3.60)  {Frozen LLM $\mathcal{M}$\\[-3pt]{\scriptsize Qwen3-8B $\cdot$ never updated}};
\node[lbl]  (y)    at (0,-4.32)  {$\hat{y}$};

\draw[ar] (x) -- (prop);
\draw[ar] (prop) -- node[lbl, right=0pt, pos=0.75] {$L^{(0)}$} (ref);
\draw[ar] (ref) -- node[lbl, right=0pt] {$\Delta$} (plus);
\draw[ar] (plus) -- node[lbl, right=0pt] {$L^{\star}$} (llm);
\draw[ar] (llm) -- (y);

\draw[ar, rounded corners=2pt]
  (0,-1.00) -- (2.35,-1.00)
  -- node[lbl, rotate=90, anchor=north, pos=0.5] {residual bypass} (2.35,-2.78) -- (plus.east);

\draw[ar, rounded corners=2pt]
  (ref.south west) -- (-2.45,-2.14) -- node[lbl, left, pos=0.5] {$\times S{\cdot}H$} (-2.45,-1.46) -- (ref.north west);

\node[lbl] (inp) at (-2.55,-3.60) {$[\,I;\,x\,]$};
\draw[ar] (inp) -- (llm.west);

\draw[-{Latex[length=1.4mm]}, densely dashed, draw=red!70]
  (llm.east) -- (3.05,-3.60) -- node[lbl, right, align=left, text=red!70, pos=0.55]
  {$\nabla\mathcal{L}_{\mathrm{CE}}$\\(activations,\\not weights)} (3.05,-0.55) -- (prop.east);
\end{tikzpicture}
\caption{\textbf{The LRT pipeline.} A task-dedicated proposer maps the question $x$ to $K$ base latents $L^{(0)}$; a recurrent refiner, re-reading $L^{(0)}$ as an external signal $u$ at every fast update, is unrolled for $S{\cdot}H$ cycles and emits a bounded residual $\Delta$; the refined latents $L^{\star}{=}L^{(0)}{+}\Delta$ are injected as soft tokens after $[\,I;x\,]$. Blue boxes are trained ($11.2$M parameters); the gray box is frozen.}
\label{fig:pipeline}
\end{figure}

SoftCoT, EBM-CoT, and LRT all inject latents through the \emph{identical} interface into the \emph{same} frozen LLM (Qwen3-8B; \citealp{yang2025qwen3}), so the proposer and refiner are the only variables and comparisons are controlled. We evaluate on two substrates. Symbolic tasks -- Countdown-4 and Sudoku -- have answer supervision but \emph{no reasoning traces}: the search that solves an instance is never recorded, so it cannot be imitated and must instead be carried out as latent computation. Natural-language tasks -- HumanEval, MBPP, and StrategyQA -- are the home turf of soft-CoT methods. The split is stark. On the natural-language tasks SoftCoT and EBM-CoT improve modestly over a zero-shot CoT baseline and LRT improves further; but on Countdown-4 both baselines \emph{collapse} -- to $5.9\%$ and $8.4\%$, far below the $30.0\%$ of zero-shot CoT -- whereas LRT reaches $56.7\%$. Off the natural-language substrate a generic proposer is not merely suboptimal but actively destructive, and only a task-dedicated proposer paired with a recurrent refiner converts latent injection back into a consistent gain.

Two scoping notes frame these numbers, and we state them here rather than in the limitations alone. First, LRT trains its proposer and refiner \emph{per task family}: it is a per-task-trained latent augmentation of a frozen decoder, not a single zero-shot general model, so its comparison class is other trained frozen-decoder modules at a matched budget. Second, the CoT baselines are prompted and training-free, and run Qwen3-8B in \emph{non-thinking} mode so that their inference budget is comparable. \S\ref{sec:reference} adds thinking mode as a labeled reference point: it beats LRT on three of the five benchmarks while spending roughly $5$--$450\times$ more inference compute. Our claim is about how far a small recurrent latent module carries a frozen decoder at a fixed, small inference cost -- not that latent reasoning dominates test-time scaling.

\section{Related Work}

\paragraph{Chain-of-thought and test-time compute.}
CoT prompting elicits multi-step reasoning by having the model emit intermediate steps \citep{wei2022chain,kojima2022large,nye2021show}, and many methods improve it by spending inference compute: self-consistency aggregates sampled traces \citep{wang2023selfconsistency}, tree- and graph-structured search organizes them \citep{yao2023tree,besta2024graph}, and compute-optimal or RL-trained reasoners scale the process further \citep{snell2024scaling,guo2025deepseek}. All of these reason in the discrete token space and lean on reasoning traces for supervision \citep{zelikman2022star,brown2020language} -- a dependence that is itself the motivation for moving reasoning off the token substrate.

\paragraph{Continuous-space reasoning.}
A complementary line moves reasoning into a model's latent space, feeding back vectors rather than decoded tokens. The most direct is \emph{hidden-state recurrence}: Coconut feeds the decoder's last hidden state back as the next input embedding, so that one continuous thought can hold a superposition of alternative continuations \citep{hao2024training}; theory shows this lets a shallow transformer explore graph-reachability problems in parallel, whereas discrete CoT must proceed sequentially \citep{zhu2025superposition}. A second family \emph{compresses} explicit CoT into dense vectors: contemplation tokens \citep{cheng2024compressed}, self-distillation into continuous space \citep{shen2025codi}, interleaved latent and text tokens \citep{su2025token}, and a latent head whose compression rate is tuned by reinforcement learning \citep{tan2025colar}; a training-free variant decodes probability-weighted mixtures of token embeddings as ``concept tokens'' \citep{zhang2025soft}. Related mechanisms add computation rather than verbalize it: implicit-CoT absorbs rationales by distillation or curricula \citep{deng2023implicit,deng2024explicit}, pause and filler tokens grant extra forward computation without emitting content \citep{goyal2024think,pfau2024lets}, and recurrent-depth pretraining unrolls layers at inference \citep{geiping2025scaling}; \citet{chen2025reasoning} and \citet{zhu2025latentsurvey} survey the area. Almost all of these pretrain or fine-tune the backbone, leaving open how to obtain useful latents when the decoder must stay fixed.

\paragraph{Latent conditioning with a frozen decoder.}
Closest to our setting are methods that keep the decoder \emph{frozen} and train only a small module supplying latents, sidestepping both the cost of fine-tuning a capable LLM and the catastrophic-forgetting risk it carries. SoftCoT runs a frozen \emph{generic} assistant LM to produce soft thoughts injected as input embeddings \citep{xu2025softcot}, and SoftCoT++ adds test-time scaling by sampling and aggregating several such thoughts \citep{xu2025softcotpp}; EBM-CoT keeps that proposer and calibrates its soft thoughts by descending a learned scalar energy field \citep{chen2025think}. Differentiable cache augmentation instead trains an offline ``coprocessor'' that writes latent embeddings into the frozen decoder's key--value cache in one forward pass \citep{liu2025deliberation}. All share LRT's frozen-decoder interface but differ in how the latents arise: each obtains them from a single feed-forward proposal, at most nudged by an energy gradient, rather than from multi-step computation unfolding through a recurrent model. LRT departs on exactly this axis, replacing the generic proposer with a task-dedicated encoder and adding a recurrent reasoner that refines the latents through many bounded residual updates before decoding.

\paragraph{Recurrent and recursive reasoning.}
Latent recurrence -- applying a shared transition repeatedly to a persistent state -- underlies Universal and Looped Transformers \citep{dehghani2019universal,giannou2023looped} and deep equilibrium models \citep{bai2019deep}, decoupling computation depth from parameter count. Recent recursive reasoners revisit it for structured tasks: HRM couples two recurrent modules at different timescales \citep{wang2025hierarchical}, TRM simplifies recursion to a single tiny network \citep{jolicoeurmartineau2025less}, GRAM makes it stochastic and multi-trajectory \citep{baek2026generative}, and EqR frames reasoning as convergence to task-conditioned attractors \citep{huang2026equilibrium}. Each is a \emph{standalone solver} trained from scratch on one task with no language ability of its own; LRT instead repurposes one (TRM) as a refiner conditioning a frozen pretrained LLM, so that iterative latent computation and broad sequence modeling come from separate components.

\paragraph{Iterative reasoning in specialized models.}
A separate line builds models that reason by iterative refinement, but outside the language-model paradigm. Energy-based reasoners cast inference as optimization over a learned energy landscape, by iterative energy minimization \citep{du2022learning}, by annealing a learned sequence of energy functions \citep{du2024learning}, or by composing separately learned energy terms \citep{du2020compositional}. Discrete-diffusion reasoners such as MGDM \citep{ye2025beyond} refine an entire candidate solution in parallel rather than left to right. These models are strong on structured symbolic problems but each is trained from scratch for a single task family on a fixed symbolic interface; extending them to open-ended natural-language reasoning is a standing challenge. LRT keeps their iterative-refinement principle while inheriting language ability for free, by attaching a recurrent reasoner to a frozen pretrained LLM rather than building a solver from scratch.

\section{Method}

\subsection{Problem Setup}
\label{sec:setup}

We target reasoning with a \emph{frozen} decoder. Let $\mathcal{M}$ be a pretrained autoregressive LLM with parameters $\theta$ and token embedding table $E\in\mathbb{R}^{|V|\times d}$ ($d{=}4096$; we use Qwen3-8B). A problem instance is a tokenized question $x$ with gold answer $y$. SoftCoT, EBM-CoT, and LRT share a common structure: a small model produces $K$ continuous latent vectors $L\in\mathbb{R}^{K\times d}$ from $x$, places them in $\mathcal{M}$'s input as soft tokens immediately after the question, and lets $\mathcal{M}$ autoregressively decode the answer. The decoder is never updated; only the modules that produce $L$ are trained, and only through the cross-entropy of $\mathcal{M}$'s output (\S\ref{sec:training}).

The three methods differ solely in how $L$ is produced. SoftCoT uses (proposer, refiner) $=$ (frozen generic assistant LM, none); EBM-CoT keeps that proposer and adds an energy-based refiner; LRT replaces both, pairing a task-dedicated encoder with a recurrent reasoner. We write the LRT pipeline as
\[
\begin{aligned}
L^{(0)} &= g_\psi(x), \qquad L^{\star} = L^{(0)} + r_\phi\big(L^{(0)}\big),\\[2pt]
\hat{y} &= \mathcal{M}\big(I,\, x,\, L^{\star}\big),
\end{aligned}
\]
where $g_\psi$ is the proposer (\S\ref{sec:proposer}), $r_\phi$ the refiner (\S\ref{sec:refiner}), and $I$ a task instruction; Figure~\ref{fig:pipeline} shows the full pipeline. Because the injection site and the frozen decoder are identical across all three methods, accuracy differences isolate the proposer and the refiner. All trainable modules operate in a small working dimension $d'{=}256$.

\subsection{Task-Dedicated Proposer}
\label{sec:proposer}

\paragraph{Motivation.}
SoftCoT obtains $L$ as the last-layer hidden states of a frozen \emph{generic} assistant LM run on $[\,I_{\text{assist}};\, x;\, \texttt{[UNK]}^{\times K}\,]$. A block of $K$ placeholder tokens appended to a prompt is off-distribution for a pretrained assistant, which therefore acts as little more than a fixed featurizer -- most severely when $x$ is short, as with the handful of integers in a Countdown instance, where the placeholders dominate. We therefore discard the generic assistant and learn a small encoder purpose-built to emit latents for $\mathcal{M}$.

\paragraph{Architecture.}
The proposer $g_\psi$ is a bidirectional Transformer encoder. It encodes the bare question $x$ -- with no task instruction, since a model trained from scratch has no instruction-following prior to invoke. The tokens of $x$ are embedded with the \emph{frozen} decoder's table $E$, so the proposer and $\mathcal{M}$ share an input geometry at no parameter cost; a trainable map $\mathrm{P}_{\!\downarrow}:\mathbb{R}^{d}\!\to\!\mathbb{R}^{d'}$ (its output scaled by $\sqrt{d'}$) reduces these embeddings to the working dimension. We append $K$ learnable query vectors and process the length-$(|x|{+}K)$ sequence with two bidirectional blocks (self-attention $+$ SwiGLU), so the queries attend to the whole problem and to one another, then read the $K$ query positions out and lift them back to the decoder's embedding space with a trainable $\mathrm{P}_{\!\uparrow}:\mathbb{R}^{d'}\!\to\!\mathbb{R}^{d}$, giving $L^{(0)}\in\mathbb{R}^{K\times d}$. Producing all $K$ latents in one bidirectional pass mirrors SoftCoT's single-pass proposal, isolating the proposer architecture as the variable under study.

\paragraph{Cost.}
The proposer adds ${\approx}\,4.2$M trainable parameters ($\mathrm{P}_{\!\downarrow}$ $1.0$M, query vectors $8$K, two blocks ${\approx}\,2.2$M, $\mathrm{P}_{\!\uparrow}$ $1.0$M) -- essentially the budget SoftCoT already spends on its assistant-to-decoder projection alone.

\subsection{Recurrent Refiner}
\label{sec:refiner}

The base latents $L^{(0)}$ are produced in a single forward pass and are not themselves the product of multi-step computation. The refiner $r_\phi$ supplies that computation: it is the recursive reasoning module of TRM \citep{jolicoeurmartineau2025less}, repurposed to iteratively improve a set of latents rather than to solve a puzzle end to end.

\paragraph{States and updates.}
The refiner maintains two states in the working dimension -- a fast scratch state $z_L$ and a slow integrating state $z_H$, both in $\mathbb{R}^{K\times d'}$ -- initialized from learned, input-independent buffers. The proposed latents enter as an external signal $u=\mathrm{P}'_{\!\downarrow}(L^{(0)})\in\mathbb{R}^{K\times d'}$, where $\mathrm{P}'_{\!\downarrow}$ is a down-projection separate from the proposer's. One high-level cycle performs $T$ fast updates followed by a single slow update (we write $T$, not $L$, to avoid collision with the latents $L$),
\[
\begin{aligned}
z_L &\leftarrow f\big(z_L,\; z_H + u\big) \qquad (\times T),\\[2pt]
z_H &\leftarrow f\big(z_H,\; z_L\big),
\end{aligned}
\]
where $f$ is a single transition block -- of the same class as the proposer's, with independent weights -- applied recursively, as in TRM. Crucially, $u$ is re-injected at \emph{every} fast update -- not only at initialization -- so refinement stays anchored to the problem while $z_H$ integrates the scratch state across cycles. After unrolling, the refined latents are formed as a bounded residual correction to the proposed base:
\[
\Delta = \mathrm{P}'_{\!\uparrow}\big(z_H\big)\in\mathbb{R}^{K\times d}, \qquad L^{\star} = L^{(0)} + \Delta .
\]
Predicting the correction $\Delta$ rather than regenerating $L^{\star}$ keeps the refiner tied to the instance-conditioned proposal; a light penalty $\lambda\lVert\Delta\rVert^2$ with $\lambda{=}0.01$ (\S\ref{sec:training}) discourages drift toward a generic, instance-agnostic latent.

\paragraph{Truncated-gradient unrolling.}
Following TRM, the refiner is unrolled deeply but trained cheaply (Algorithm~\ref{alg:lrt}). We run $S$ outer iterations of $H$ high-level cycles each; every cycle but the last runs under stop-gradient, advancing $z_L,z_H$ without building a computation graph, so backpropagation traverses only the final cycle while the preceding unroll merely warms up the latent state. Unlike TRM's adaptive-computation variant, which supervises every step, we take the cross-entropy \emph{once}, on the final $L^{\star}$. Gradients flow from $\mathcal{M}$'s answer logits -- through the frozen decoder, whose activations are differentiated but whose weights are untouched -- into $L^{\star}$, $\Delta$, the refiner and -- in Stage~1 -- the proposer.

\paragraph{Cost.}
The refiner adds ${\approx}\,7$M trainable parameters, comparable to EBM-CoT's energy module. LRT's two modules together total ${\approx}\,11$M trainable parameters; the $8$B-parameter decoder is frozen throughout, so LRT trains fewer than $0.2\%$ of the parameters it reasons with.

\begin{algorithm}[t]
\caption{LRT forward pass}
\label{alg:lrt}
\small
\begin{algorithmic}[1]
\Require problem $x$; frozen decoder $\mathcal{M}$; proposer $g_\psi$; refiner $r_\phi$ with learned buffers $z_L^{0}, z_H^{0}$
\State $L^{(0)} \gets g_\psi(x)$ \Comment{$K$ base latents (\S\ref{sec:proposer})}
\State $u \gets \mathrm{P}'_{\!\downarrow}(L^{(0)})$;\quad $z_L, z_H \gets z_L^{0}, z_H^{0}$
\For{$s = 1$ \textbf{to} $S$}
  \For{$h = 1$ \textbf{to} $H$}
    \State $\textit{grad} \gets (s{=}S \,\wedge\, h{=}H)$ \Comment{else wrap updates in $\mathrm{sg}[\cdot]$}
    \For{$t = 1$ \textbf{to} $T$}
      \State $z_L \gets f(z_L,\; z_H + u)$
    \EndFor
    \State $z_H \gets f(z_H,\; z_L)$
  \EndFor
\EndFor
\State $\Delta \gets \mathrm{P}'_{\!\uparrow}(z_H)$;\quad $L^{\star} \gets L^{(0)} + \Delta$ \Comment{residual}
\State \Return $\hat{y} \gets \mathcal{M}(I, x, L^{\star})$
\end{algorithmic}
\end{algorithm}

\subsection{Injection and Decoding}
\label{sec:injection}

Following SoftCoT, $\mathcal{M}$ receives the concatenation $[\,I;\, x;\, L^{\star}\,]$: a task instruction $I$ and the question $x$ mapped to input embeddings through $E$, followed by the $K$ refined latents as additional input embeddings. Because $\mathrm{P}'_{\!\uparrow}$ and the residual already output in $\mathbb{R}^{d}$, $L^{\star}$ needs no further projection. The instruction supplies the decoder's instruction-following prior and $L^{\star}$ the instance-specific reasoning; $\theta$ is never modified.

\subsection{Training and Inference}
\label{sec:training}

\paragraph{Two-stage training.}
LRT is trained in two stages, each optimizing a single module against the cross-entropy of the frozen decoder; gradients pass through $\mathcal{M}$'s activations but never update $\theta$. \emph{Stage 1} trains the proposer alone: the base latents $L^{(0)}$ are injected directly, as $[I;x;L^{(0)}]$, and $g_\psi$ is optimized so that $\mathcal{M}$ decodes the correct answer. \emph{Stage 2} freezes $g_\psi$ and trains the refiner on
\[
\mathcal{L} = \mathcal{L}_{\text{CE}}\big(\mathcal{M}(I,x,L^{\star}),\, y\big) \;+\; \lambda\,\lVert\Delta\rVert^{2},
\]
the answer cross-entropy under the refined latents plus the residual penalty of \S\ref{sec:refiner}, with $\lambda{=}0.01$. Since $g_\psi$ is frozen, each instance's base latents $L^{(0)}$ are precomputed once and cached, so Stage~2 needs no forward pass through the proposer. Across both stages the only supervision is the final answer -- or, for StrategyQA, the answer with its reference rationale -- and never a record of the search that produced it.

\paragraph{Refiner unrolling.}
We unroll the refiner for $S{=}3$ outer iterations of $H{=}3$ high-level cycles; each cycle applies $T{=}4$ fast updates and one slow update -- five passes through the transition block -- for $3{\times}3{\times}5{=}45$ passes in total. As described in \S\ref{sec:refiner}, only the final cycle (five passes) is differentiated; the preceding $40$ run under stop-gradient, and the cross-entropy is taken once, on $L^{\star}$.

\paragraph{Inference.}
At test time the proposer emits its query latents and we keep the first $K_{\text{infer}}{=}4$ before the refiner -- a train/inference asymmetry ($K{=}32$ during training) adopted from EBM-CoT, whose latent-count ablation peaks near four. The refiner then runs the full $S{\times}H$ recursion (no stop-gradient is needed without a backward pass), and $\mathcal{M}$ greedy-decodes the answer from $[I;x;L^{\star}]$. Optimization hyperparameters are reported with the experiments.

\section{Experiments}

\subsection{Experimental Setup}
\label{sec:exp_setup}

\paragraph{Benchmarks.}
We evaluate on two families of reasoning tasks. The \emph{symbolic} family stresses computation away from the natural-language substrate: \textbf{Countdown-4} (CD4), in which four integers must be combined with arithmetic operations to reach a target value \citep{gandhi2024stream}, and \textbf{Sudoku}, $9{\times}9$ constraint-satisfaction puzzles \citep{ye2025beyond}. Both supply answer supervision but no ground-truth search traces. The \emph{natural-language} family comprises \textbf{HumanEval} \citep{chen2021evaluating} and \textbf{MBPP} \citep{austin2021program} -- Python function synthesis from docstrings -- and \textbf{StrategyQA} \citep{geva2021strategyqa}, yes/no questions requiring implicit multi-step reasoning.

\paragraph{Metrics.}
We report exact solve rate on Countdown-4 and Sudoku, pass@1 on HumanEval and MBPP, and accuracy on StrategyQA. All values are percentages. Table~\ref{tab:main} reports mean\,$\pm$\,standard deviation over three random seeds; the ablation tables report single-seed values. \textbf{Avg.}\ is an unweighted mean, over the five benchmarks unless a caption states otherwise.

\paragraph{Backbone and baselines.}
Every LLM-based method uses the same frozen Qwen3-8B decoder \citep{yang2025qwen3}. Table~\ref{tab:main} compares two groups under this single controlled setting. The first adds no trainable parameters: \emph{Zero-Shot CoT} \citep{kojima2022large} (henceforth zero-shot CoT) and a \emph{Direct} (no CoT) baseline that answers without intermediate steps. The second is the family of \emph{trained frozen-decoder latent modules} to which LRT belongs: \emph{SoftCoT} \citep{xu2025softcot} and \emph{EBM-CoT} \citep{chen2025think},\footnote{EBM-CoT has no official implementation; the EBM-CoT results reported here are from our own reproduction of the method as described by \citet{chen2025think} and may deviate from the authors' intended setup.} which -- like LRT -- keep the decoder frozen and train only a small latent module. All of these share LRT's interface exactly -- same decoder, same prompt, same training data and budget -- so the latent module is the only variable. Appendix~\ref{app:peft} adds two parameter-efficient prompt-adaptation controls at a matched ${\approx}11$M trainable budget, \emph{Prefix-tuning} \citep{li2021prefix} and \emph{P-tuning~v2} \citep{liu2022ptuningv2}, which test whether task supervision plus trainable soft prompts alone reproduce LRT's gain; they do not.

\paragraph{What is deliberately \emph{not} a baseline, and why.}
Three families are excluded by design rather than oversight. (i) \emph{Methods that update decoder weights} -- Coconut \citep{hao2024training}, CODI \citep{shen2025codi}, Token-Assorted \citep{su2025token}, and adapter- or LoRA-style tuning \citep{houlsby2019parameter,hu2022lora} -- leave the frozen-decoder setting that defines our study, and fine-tuning is known to degrade strong instruction-tuned backbones \citep{xu2025softcot}. (ii) \emph{Test-time scaling} such as self-consistency \citep{wang2023selfconsistency} and SoftCoT++ \citep{xu2025softcotpp} is orthogonal: it aggregates predictions from whatever base method is used, so it composes with LRT rather than competing. (iii) \emph{From-scratch symbolic solvers} (MGDM, TRM, EqR) cannot process natural language at all, and appear in Table~\ref{tab:reference} as reference points. Comparisons crossing these boundaries are informative but uncontrolled, and labeled as such throughout.

\paragraph{Reproduction and harness notes.}
Three details affect how the baseline numbers should be read; Appendix~\ref{app:reprod} states them in full. EBM-CoT has no official implementation, so its rows are our reproduction, validated against the authors' reported StrategyQA result ($70.96$ vs.\ $71.04$ ours). The CoT baselines run Qwen3 in non-thinking mode for budget comparability, which departs substantially from the official technical-report numbers; we validated the harness by reproducing the official MBPP result with thinking mode on. TRM is not natively a sequence-to-sequence model, so Countdown-4 required an adaptation of our own, whose limits we state in \S\ref{sec:reference}.

\paragraph{Training.}
LRT is trained in the two stages of \S\ref{sec:training}; hyperparameters are in Appendix~\ref{app:impl}.

\begin{table*}[t]
\centering
\small
\setlength{\tabcolsep}{6pt}
\begin{tabular}{l cc ccc c}
\toprule
& \multicolumn{2}{c}{\textbf{Symbolic}} & \multicolumn{3}{c}{\textbf{Natural Language}} & \\
\cmidrule(lr){2-3}\cmidrule(lr){4-6}
\textbf{Method} & CD4 & Sudoku & HumanEval & MBPP & StrategyQA & \textbf{Avg.} \\
\midrule
\multicolumn{7}{l}{\textit{Frozen decoder, no trainable parameters}}\\
\quad Direct (no CoT)        & 27.8\,$\pm$\,1.6 & 17.3\,$\pm$\,0.3 & 13.4\,$\pm$\,1.1 & 28.7\,$\pm$\,1.8 & 67.4\,$\pm$\,6.1 & 30.9 \\
\quad Zero-Shot CoT          & 30.0\,$\pm$\,0.6 & 23.9\,$\pm$\,0.2 & 15.9\,$\pm$\,0.5 & 35.4\,$\pm$\,1.6 & 69.9\,$\pm$\,1.4 & 35.0 \\
\midrule
\multicolumn{7}{l}{\textit{Propose-then-inject (frozen decoder, trained latent module)}}\\
\quad SoftCoT \citep{xu2025softcot}                & 5.9\,$\pm$\,0.2 & 10.4\,$\pm$\,0.1 & 20.7\,$\pm$\,0.9 & 40.2\,$\pm$\,2.1 & 70.2\,$\pm$\,6.5 & 29.5 \\
\quad EBM-CoT \citep{chen2025think}                & 8.4\,$\pm$\,0.1 & 17.2\,$\pm$\,0.4 & 25.0\,$\pm$\,1.6 & 46.1\,$\pm$\,3.6 & 71.0\,$\pm$\,5.7 & 33.5 \\
\quad \textbf{LRT (Ours)}    & \textbf{56.7\,$\pm$\,1.9} & \textbf{49.2\,$\pm$\,1.5} & \textbf{37.8\,$\pm$\,3.3} & \textbf{51.5\,$\pm$\,1.7} & \textbf{75.1\,$\pm$\,2.3} & \textbf{54.1} \\
\bottomrule
\end{tabular}
\caption{Main results: a single controlled comparison in which every method uses the \emph{same} frozen Qwen3-8B decoder, prompts, training data, and budget, so that the trained latent module is the only variable. Countdown-4 and Sudoku report exact solve rate, HumanEval and MBPP pass@1, StrategyQA accuracy; entries are mean\,$\pm$\,standard deviation in percent over three seeds, and \textbf{Avg.}\ is the unweighted mean over the five benchmarks. The prompted baselines run in non-thinking mode (\S\ref{sec:exp_setup}). Methods outside this setting -- thinking-mode prompting, from-scratch symbolic solvers, and parameter-efficient prompt adaptation -- appear as reference points in Table~\ref{tab:reference} and Appendix~\ref{app:peft}. \textbf{Bold}: best in each column.}
\label{tab:main}
\end{table*}

\begin{table}[t]
\centering
\small
\setlength{\tabcolsep}{2.9pt}
\begin{tabular}{l ccccc}
\toprule
 & CD4 & Sudoku & HEval & MBPP & SQA \\
\midrule
\textbf{LRT (Ours)} & 56.7 & 49.2 & 37.8 & \textbf{51.5} & 75.1 \\
\midrule
\multicolumn{6}{l}{\textit{Same backbone, thinking mode (not compute-matched)}}\\
\quad Qwen3-8B      & \textbf{85.3} & 0.5$^{\dagger}$ & \textbf{92.1} & 49.3$^{\ddagger}$ & \textbf{76.4} \\[1pt]
\quad {\footnotesize TFLOP/example} & {\footnotesize 144.9} & {\footnotesize 453.9} & {\footnotesize 63.4} & {\footnotesize 67.3} & {\footnotesize 8.9} \\
\midrule
\multicolumn{6}{l}{\textit{From-scratch symbolic solvers (no language ability)}}\\
\quad MGDM 6M       & 52.0 & 96.9 & \multicolumn{3}{c}{\emph{undefined}} \\
\quad TRM 7M        & 1.2  & 97.2 & \multicolumn{3}{c}{\emph{undefined}} \\
\quad EqR 7M        & --   & \textbf{99.8} & \multicolumn{3}{c}{\emph{undefined}} \\
\bottomrule
\end{tabular}
\caption{\textbf{Reference points outside the controlled setting of Table~\ref{tab:main}.} Thinking mode is \emph{not} compute-matched: at ${\approx}1$ TFLOP per example, LRT spends roughly $9\times$ (StrategyQA) to $450\times$ (Sudoku) less. Thinking-mode rows use fixed random test subsets ($n{=}300/200/300$ for Countdown-4/Sudoku/MBPP); all other rows use the full evaluation sets. \textbf{Bold}: best in each column; a dash marks a number the original work does not report. MGDM \citep{ye2025beyond}, TRM \citep{jolicoeurmartineau2025less}, and EqR \citep{huang2026equilibrium} are trained from scratch on one symbolic task and have no language interface, so their natural-language entries are undefined rather than missing. $^{\dagger}$Not a truncation artifact; $^{\ddagger}$name-normalized (Appendix~\ref{app:reprod}).}
\label{tab:reference}
\end{table}

\subsection{Main Results}
\label{sec:main_results}

Table~\ref{tab:main} reports results across all five benchmarks.

\paragraph{Symbolic reasoning.}
The symbolic benchmarks expose a sharp failure of the existing latent-injection methods. On Countdown-4, SoftCoT and EBM-CoT reach only $5.9$ and $8.4$ -- far below both the no-reasoning Direct baseline ($27.8$) and zero-shot CoT ($30.0$) -- and the same ordering recurs on Sudoku. Injecting latents from a generic assistant is thus not merely unhelpful off the natural-language substrate but actively destructive, corroborating the diagnosis of \S\ref{sec:proposer}. LRT reverses this: with a task-dedicated proposer and a recurrent refiner it reaches $56.7$ on Countdown-4 and $49.2$ on Sudoku, the only frozen-decoder method to improve over zero-shot CoT on either task, and by a wide margin ($+26.7$ and $+25.3$ points). On Countdown-4 it even surpasses MGDM ($52.0$; Table~\ref{tab:reference}), a diffusion model trained from scratch specifically for symbolic decoding.

\paragraph{Natural-language reasoning.}
On HumanEval, MBPP, and StrategyQA -- the substrate for which soft-thought methods were designed -- latent injection behaves as intended: SoftCoT and EBM-CoT both improve over Direct prompting and zero-shot CoT. LRT is strongest on all three (e.g.\ $37.8$ vs.\ $25.0$ for EBM-CoT on HumanEval), showing that a task-dedicated proposer and a recurrent refiner help even where the generic pipeline already works. Averaged over the five benchmarks, LRT reaches $54.1$, against $33.5$ for EBM-CoT and $35.0$ for zero-shot CoT; notably, SoftCoT's average ($29.5$) falls below the no-reasoning Direct baseline ($30.9$), as its symbolic collapse cancels its natural-language gains. Empirically, we observed that both SoftCoT and EBM-CoT collapse to predicting a near-uniform latent across instances, which may in part explain the sharp drop in performance. Note that injecting a constant latent is \emph{not} the same as direct decoding, since those vectors still play a non-trivial part in the model's attention patterns.

\subsection{Reference Points Outside the Controlled Setting}
\label{sec:reference}

Table~\ref{tab:main} holds the decoder, prompt, data, and budget fixed. Three comparisons that break one of those controls are nonetheless informative, and we report them separately so that they are not mistaken for baselines.

\paragraph{Thinking-mode prompting.}
In thinking mode with Qwen3's recommended sampling and a $32{,}768$-token budget, the same backbone reaches $85.3$ on Countdown-4 and $92.1$ on HumanEval, well above LRT; LRT leads on Sudoku ($49.2$ vs.\ $0.5$) and MBPP ($51.5$ vs.\ $49.3$), and thinking mode edges ahead on StrategyQA ($76.4$ vs.\ $75.1\pm2.3$, within one standard deviation) -- while spending $8.9$--$453.9$ TFLOP per example against LRT's ${\approx}1$--$2$. Two things follow. Our claim must be read as scoped: LRT outperforms \emph{prompted, non-thinking} CoT at a fraction of its cost, not test-time scaling in general. But a weak prompting setup cannot explain the main result either, since SoftCoT and EBM-CoT use the same decoder, prompts, data, and budget as LRT and reach $5.9$ and $8.4$ on Countdown-4. It is also notable that the task where thinking mode spends the most compute, Sudoku, is exactly where it fails ($0.5$) while a $7$M recurrent refiner reaches $49.2$: token-space search is not a substitute for iterative computation on a constraint-satisfaction substrate.

\paragraph{From-scratch symbolic solvers.}
MGDM and TRM solve Sudoku almost perfectly ($96.9$, $97.2$) and EqR essentially completely ($99.8$), far above LRT's $49.2$. We state this plainly: on Sudoku a specialized solver is much better than LRT, and the frozen-decoder route trades peak symbolic accuracy for one method that also handles language. Two observations qualify the comparison rather than rescue it. First, these models do not transfer: trained from scratch for symbolic decoding, they have no language interface at all. Second, their competence is uneven: TRM, the very reasoner LRT adopts as its refiner, scores $1.2$ on Countdown-4 despite near-solving Sudoku. We attribute this partly to task structure: Countdown-4 admits an unusually sequential solution, a chain of arithmetic operations applied in order, which a from-scratch model with no sequence-modeling prior struggles to capture, whereas Sudoku suits TRM's grid recurrence. The honest caveat is that TRM does not natively support sequence-to-sequence tasks; our adaptation (Appendix~\ref{app:reprod}) is simple and stronger ones may exist. The central point survives: without \emph{some} sequential prior, TRM struggles on a fundamentally sequential task, and LRT is built around exactly this division of labor.

\paragraph{Parameter-efficient prompt adaptation.}
At a trainable budget matched to LRT's ${\approx}11$M, prefix-tuning reaches $12.5$ and P-tuning~v2 $28.4$ on Countdown-4 (Appendix~\ref{app:peft}), the latter merely matching the no-latent Direct baseline ($27.8$): task supervision plus trainable soft prompts do not by themselves produce the gain. Making the injected vectors instance-conditioned adds $+13.6$ (to $42.0$), and recurrent refinement a further $+14.7$ (to $56.7$).

\subsection{Scaling the Recurrent Refiner}
\label{sec:scaling}

In LRT the decoder is frozen, so all learned computation lives in the proposer and the refiner -- and of the two, the refiner is where the iterative reasoning happens, making its capacity a natural lever. Table~\ref{tab:main} uses a $7$M refiner, chosen for parity with EBM-CoT's energy module (\S\ref{sec:refiner}); we now enlarge it while \emph{everything else is held fixed}.

\begin{figure}[t]
\centering
\includegraphics[width=0.8\columnwidth]{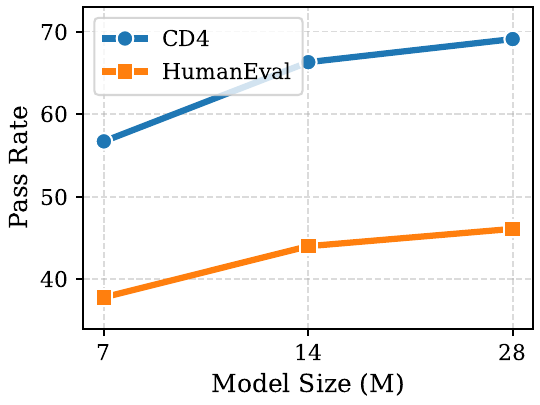}
\caption{\textbf{Scaling the recurrent refiner.} Pass rate on Countdown-4 and HumanEval as the TRM refiner grows from $7$M to $28$M parameters, with the frozen Qwen3-8B decoder and the proposer held fixed (metrics as in Table~\ref{tab:main}). Both tasks improve monotonically, with the largest gain at the first doubling; the $7$M point is the configuration reported in Table~\ref{tab:main}.}
\label{fig:scaling}
\end{figure}

Figure~\ref{fig:scaling} scales the TRM refiner across $7$M, $14$M, and $28$M parameters and measures Countdown-4 and HumanEval. Both improve monotonically: Countdown-4 rises from $56.7$ to $66.3$ to $69.1$, and HumanEval from $37.8$ to $44.0$ to $46.1$. The gains are largest at the first doubling ($+9.6$ and $+6.2$ points) and taper at $28$M ($+2.8$ and $+2.1$). Crucially, the improvement is bought entirely with refiner capacity: the $8$B decoder never changes. Reasoning quality thus scales with the recurrent reasoner that carries the computation, separately from the LLM that supplies sequence modeling, and the $7$M setting used elsewhere is a conservative, parameter-matched operating point with clear headroom.

\subsection{Mechanistic Ablations}
\label{sec:ablations}

The comparison in \S\ref{sec:main_results} varies two things at once: relative to LRT, SoftCoT and EBM-CoT each swap \emph{both} the proposer (task-dedicated $\rightarrow$ generic) and the refiner. We now isolate the two components.

\paragraph{Factorial decomposition.}
LRT pairs a proposer and a refiner (\S\ref{sec:proposer}, \S\ref{sec:refiner}), so the natural ablation is the $2{\times}3$ grid over proposer $\in\{$generic assistant, task-dedicated$\}$ and refiner $\in\{$none, energy, recurrent$\}$. SoftCoT, EBM-CoT, and LRT are three of its six cells; Table~\ref{tab:factorial} fills the rest, and two of the three new cells correspond to alternative pairings a reader might expect: \emph{task-dedicated\,+\,none} is SoftCoT with our proposer substituted, and \emph{task-dedicated\,+\,energy} is EBM-CoT with our proposer substituted. All six cells are reported. Holding the refiner at \emph{none}, swapping the generic assistant for a task-dedicated proposer lifts the average from $12.3$ to $37.2$; holding the proposer task-dedicated, the recurrent refiner contributes roughly three times the energy refiner's gain ($+10.7$ vs.\ $+3.4$). The two also interact -- the recurrent refiner adds $+10.7$ on a task-dedicated proposer but only $+9.3$ on a generic one, and on Countdown-4 every generic-proposer cell stays below the Direct baseline ($27.8$) whatever the refiner. Neither component alone accounts for LRT: the proposer must place the base latents on a usable manifold before the refiner can compute over them.

\begin{table}[t]
\centering
\small
\setlength{\tabcolsep}{4.5pt}
\begin{tabular}{l cccc}
\toprule
 & CD4 & Sudoku & HEval & \textbf{Avg.} \\
\midrule
\multicolumn{5}{l}{\textit{Generic proposer}}\\
\quad + no refiner        & 5.9  & 10.4 & 20.7 & 12.3 \\
\quad + energy refiner    & 8.4  & 17.2 & 25.0 & 16.9 \\
\quad + recurrent refiner & 14.8 & 20.6 & 29.4 & 21.6 \\
\midrule
\multicolumn{5}{l}{\textit{Task-dedicated proposer}}\\
\quad + no refiner        & 42.0 & 38.5 & 31.2 & 37.2 \\
\quad + energy refiner    & 46.5 & 41.8 & 33.4 & 40.6 \\
\quad + recurrent refiner & \textbf{56.7} & \textbf{49.2} & \textbf{37.8} & \textbf{47.9} \\
\bottomrule
\end{tabular}
\caption{\textbf{Factorial ablation} over proposer and refiner. \textit{Generic\,+\,no refiner} is SoftCoT, \textit{generic\,+\,energy} is EBM-CoT, and \textit{task-dedicated\,+\,recurrent} is LRT; the other three rows are new ablations. Here \textbf{Avg.}\ is the mean of the three columns shown, \emph{not} the five-benchmark average of Table~\ref{tab:main}.}
\label{tab:factorial}
\end{table}

\paragraph{Depth versus capacity.}
\S\ref{sec:scaling} scaled the refiner by \emph{parameters}; we now vary \emph{depth} at fixed parameters (Appendix~\ref{app:depth}). A non-recurrent refiner with the \emph{same} $7$M budget but a single feed-forward pass reaches only $47.5/33.0$ on Countdown-4/HumanEval against LRT's $56.7/37.8$; and a refiner trained at $S{=}H{=}3$ but unrolled deeper \emph{at test time only} keeps improving, indicating a convergent iterative process rather than a fixed-length circuit.

\paragraph{Refinement is incremental.}
Decoding the answer after each high-level cycle (Appendix~\ref{app:mech}) shows accuracy climbing monotonically across the unroll rather than appearing only at the final step, while the residual $\lVert L^{(t)}-L^{(t-1)}\rVert$ contracts toward a fixed point: the refiner performs incremental, convergent computation, not one-shot calibration. Appendix~\ref{app:refiner-design} confirms each design choice earns its place -- regenerating $L^{\star}$ instead of the residual $\Delta$, or injecting the base only at initialization, each costs $3$--$4$ average points.

\subsection{Where Does the Computation Happen?}
\label{sec:where}

LRT combines a recurrent reasoner with a frozen LLM, inviting a natural worry: is the gain simply TRM, with the LLM reduced to a transcription device? Four measurements bound the answer from both sides.

\emph{TRM alone does not do the work.} As a standalone solver it scores $1.2$ on Countdown-4 against LRT's $56.7$ and cannot address the natural-language benchmarks at all (Table~\ref{tab:reference}): on four of five benchmarks TRM alone is not merely worse but inapplicable, Sudoku being the exception noted in \S\ref{sec:reference}. \emph{The refiner needs on-manifold latents.} The same refiner applied to latents from a \emph{generic} proposer reaches only $14.8$ on Countdown-4 (Table~\ref{tab:factorial}), below the no-latent Direct baseline of $27.8$: recurrence is not a free-standing engine, and anchoring it in the decoder's representation space is what makes it useful. \emph{The decoder is load-bearing.} Swapping it changes results meaningfully -- $54.1$, $51.8$, $49.0$ average for Qwen3-8B, Llama-3.1-8B, Qwen3-4B (Appendix~\ref{app:decoders}) -- and removing the instruction $I$ at inference costs $3.4$ points (Appendix~\ref{app:protocol}); neither should matter much if the decoder merely transcribed a finished answer, since any of these backbones can verbalize one. \emph{The answer is not sitting in the latent.} A linear probe on cached $L^{\star}$ predicting whether the decoder will solve the instance reaches $57.5\%$ against a $52.5\%$ majority baseline, with unrefined latents at chance (Appendix~\ref{app:mech}); even the outcome, let alone the answer string, is only weakly linearly readable. Nor can the answer have been memorized: Countdown-4 evaluation holds out targets.

This evidence is convergent but correlational: a linear probe bounds only \emph{linearly} decodable information, and the decoder-swap argument presumes comparable verbalization ability across backbones. We claim the computation is distributed across refiner and decoder rather than concentrated in either, not that we have measured the split.

\section{Conclusion}
\label{sec:conclusion}

We presented Latent Recurrent Thoughts (LRT), which reasons with a frozen LLM by pairing a task-dedicated proposer with a TRM-based recurrent refiner, training fewer than $0.2\%$ of the parameters involved: the proposer maps a problem to base latents the frozen decoder can use, and the refiner computes over them, emitting a bounded residual correction. Across symbolic and natural-language reasoning LRT improves substantially over prior frozen-decoder latent-injection methods under an identical decoder, prompt, data, and budget, and over budget-matched zero-shot CoT; neither component suffices alone, and the computation is distributed across refiner and decoder rather than carried by either (\S\ref{sec:where}). The result is scoped -- modules are trained per task family, and large test-time budgets remain stronger on some tasks -- but within it, a small recurrent module in front of a frozen LLM supplies much of the computation chain-of-thought otherwise externalizes.

\section*{Limitations}

\paragraph{Per-task modules, and what that costs the claim.}
LRT keeps the decoder frozen but trains a proposer and a refiner against it, per task family. It is therefore \emph{not} a single general model applied zero-shot: each new task requires training new modules, and the method is best described as per-task-trained latent augmentation of a frozen decoder rather than a general reasoning method. This bounds the fair comparison class: the controlled comparison in Table~\ref{tab:main} is against other trained frozen-decoder latent modules, and comparisons to prompting are reported at matched inference budget, with thinking-mode prompting reported separately as an unmatched reference point (\S\ref{sec:reference}). Training additionally relies on answer supervision and, for the symbolic benchmarks, on the ability to generate synthetic training sets -- though Appendix~\ref{app:peft} shows $1{,}000$ Countdown-4 examples already reach $52.7$, so the requirement is for \emph{some} answer-labeled data rather than a great deal of it. Tasks where even that is unavailable remain out of reach.

\paragraph{Generalization is within-task, not across tasks.}
Our Countdown-4 evaluation holds out targets, so the reported accuracy does reflect systematic generalization to unseen instances rather than memorization. But we have not tested transfer \emph{across} tasks or difficulty levels: whether modules trained on Countdown-4 help on harder Countdown variants, or Sudoku modules on Kakuro, is untested and we make no claim about it. The per-task protocol matches SoftCoT and EBM-CoT, which makes the comparison controlled but leaves cross-task generality an open question for all three.

\paragraph{Scope of evaluation.}
Our experiments use a single $8$B-parameter decoder and five benchmarks. Appendix~\ref{app:decoders} varies the decoder, but all backbones are of comparable scale; whether the division of labor between a frozen decoder and a small recurrent reasoner holds at substantially larger scales, or across broader task distributions, remains open. The parameter-efficient controls of Appendix~\ref{app:peft} were run on Countdown-4 only.

\paragraph{Method caveats.}
On Sudoku, symbolic-specialized solvers remain far ahead of LRT (Table~\ref{tab:reference}); the frozen-decoder route trades peak symbolic accuracy for a single method that also handles natural language. The refiner's recurrent unroll adds inference compute that grows with depth (Appendices~\ref{app:depth} and~\ref{app:compute}). Our mechanistic analyses (\S\ref{sec:ablations}, \S\ref{sec:where}, Appendix~\ref{app:mech}) are correlational: the monotone intermediate-decoding curve, the contracting latent residual, and the weakly-above-baseline linear probe are evidence about where computation happens, not proof, and the probe in particular bounds only linear decodability.

\newpage

\bibliography{custom}

\appendix

\section{Implementation and Training Details}
\label{app:impl}

\paragraph{Architecture.}
The proposer is a bidirectional encoder: an input projection $\mathrm{P}_{\!\downarrow}$ maps decoder embeddings of $x$ to working width $d'{=}256$, two pre-norm transformer blocks attend over $K{=}32$ learned query vectors appended to the sequence, and an output projection $\mathrm{P}_{\!\uparrow}$ returns the $K$ base latents to decoder width $d$. The refiner reuses one transition block of the same class (with independent weights) as $f$, together with projections $\mathrm{P}'_{\!\downarrow},\mathrm{P}'_{\!\uparrow}$ and learned initial buffers $z_L^{0},z_H^{0}$. Table~\ref{tab:params} breaks down the trainable parameters; the Qwen3-8B decoder is frozen throughout, so LRT trains $11.2$M parameters, about $0.14\%$ of the model it reasons with.

\begin{table}[t]
\centering
\small
\begin{tabular}{l r}
\toprule
Component & Params \\
\midrule
\textit{Proposer} & \\
\quad input projection $\mathrm{P}_{\!\downarrow}$ & 1.0M \\
\quad query embeddings ($K{=}32$) & 0.008M \\
\quad 2 encoder blocks & 2.2M \\
\quad output projection $\mathrm{P}_{\!\uparrow}$ & 1.0M \\
\textit{Refiner} & \\
\quad input projection $\mathrm{P}'_{\!\downarrow}$ & 1.0M \\
\quad transition block $f$ & 4.8M \\
\quad output projection $\mathrm{P}'_{\!\uparrow}$ & 1.0M \\
\quad initial buffers $z_L^{0},z_H^{0}$ & 0.2M \\
\midrule
Total trainable & 11.2M \\
Frozen decoder (Qwen3-8B) & 8.2B \\
\bottomrule
\end{tabular}
\caption{Trainable parameter breakdown; counts are approximate at working width $d'{=}256$.}
\label{tab:params}
\end{table}

\paragraph{Optimization.}
Both stages use AdamW ($\beta_1{=}0.9$, $\beta_2{=}0.999$, $\epsilon{=}10^{-8}$) with a cosine learning-rate schedule, $5\%$ linear warmup, weight decay $0.01$, and gradient clipping at $1.0$, in \texttt{bfloat16}. Stage~1 trains the proposer at peak learning rate $3{\times}10^{-4}$; Stage~2 trains the refiner at $2{\times}10^{-4}$. Both run for $30$ epochs at batch size $64$ on a single $96$\,GB GPU, with gradient checkpointing. Table~\ref{tab:hparams} collects the hyperparameters.

\begin{table}[t]
\centering
\small
\begin{tabular}{l l}
\toprule
Hyperparameter & Value \\
\midrule
Optimizer & AdamW \\
Weight decay & $0.01$ \\
LR schedule & cosine, $5\%$ warmup \\
Peak LR (Stage~1 / Stage~2) & $3{\times}10^{-4}$ / $2{\times}10^{-4}$ \\
Gradient clipping & $1.0$ \\
Batch size & $64$ \\
Epochs per stage & $30$ \\
Precision & \texttt{bfloat16} \\
Working width $d'$ & $256$ \\
Latents $K$ / $K_{\text{infer}}$ & $32$ / $4$ \\
Refiner unroll $S,H,T$ & $3,3,4$ \\
Residual penalty $\lambda$ & $0.01$ \\
Hardware & $1{\times}96$\,GB GPU \\
\bottomrule
\end{tabular}
\caption{LRT hyperparameters used for all reported results.}
\label{tab:hparams}
\end{table}

\paragraph{Datasets.}
Table~\ref{tab:data} lists sizes and splits. Countdown-4 instances are generated following \citet{gandhi2024stream} and Sudoku puzzles following \citet{ye2025beyond}; HumanEval, MBPP, and StrategyQA use their public splits. Countdown-4 evaluation \emph{holds out targets}: no target value in the test set appears in training, so the reported accuracy reflects generalization to unseen targets rather than fitting the synthetic distribution.

\paragraph{What the supervision consists of.}
The claim that LRT trains without reasoning traces is scoped to the symbolic benchmarks, and we state the full picture here. Countdown-4 and Sudoku are supervised strictly by the final answer; the search that produces it is never recorded and never used. StrategyQA is the one exception: in the main results its target text is the answer together with the dataset's reference rationale, following prior work. Evaluation is unaffected -- StrategyQA is scored by exact match on the normalized yes/no answer only, never on the rationale. To remove any ambiguity, we also trained StrategyQA with \emph{answer-only} supervision: it reaches $76.9$, against $75.1\pm2.3$ for the rationale-supervised setting reported in Table~\ref{tab:main}. The rationale is therefore not necessary for LRT's result, and no benchmark in this paper depends on trace supervision.

\begin{table}[t]
\centering
\small
\begin{tabular}{l l r r}
\toprule
Dataset & Source & Train & Eval \\
\midrule
Countdown-4 & generated & 100{,}000 & 1{,}000 \\
Sudoku      & generated & 100{,}000 & 1{,}000 \\
HumanEval   & public    & --        & 164 \\
MBPP        & public    & 374       & 500 \\
StrategyQA  & public    & 2{,}061   & 229 \\
\bottomrule
\end{tabular}
\caption{Dataset sizes. HumanEval is evaluation-only; the code modules are trained on the MBPP training split.}
\label{tab:data}
\end{table}

\paragraph{Inference.}
At test time the proposer emits its $K{=}32$ query latents and the first $K_{\text{infer}}{=}4$ are kept (Table~\ref{tab:kd}); the refiner runs the full $S{\times}H$ recursion with no stop-gradient, and the decoder greedy-decodes the answer from $[I;x;L^{\star}]$.

\section{Reproduction and Harness Notes}
\label{app:reprod}

\paragraph{Baseline reproduction.}
SoftCoT is reproduced from the authors' released code; EBM-CoT, which has no public implementation, is reproduced from the description of \citet{chen2025think}. Both use the same frozen Qwen3-8B decoder, the same assistant model for their generic proposer, and the same stage-wise training budget as LRT, so differences reflect the latent module rather than the backbone or compute. To check the faithfulness of the EBM-CoT reproduction we compare on StrategyQA, the one benchmark shared with the original paper: $70.96$ reported by \citet{chen2025think} against $71.04$ for our reproduction. Reproduced numbers may nonetheless deviate from the authors' intended setup on the benchmarks they did not report.

\paragraph{Thinking versus non-thinking mode.}
Qwen3 exposes a ``thinking'' mode that emits a long internal trace before answering. All rows of Table~\ref{tab:main} use \emph{non-thinking} mode so that the prompted baselines and the latent methods spend comparable inference compute; this departs substantially from the official technical-report numbers, especially on the coding tasks. To verify that the deviation reflects the mode rather than a harness bug, we re-ran with thinking mode enabled and reproduced the official MBPP figure. Those runs appear as reference points in Table~\ref{tab:reference} and are discussed in \S\ref{sec:reference}.

Thinking-mode runs use Qwen3's recommended sampling with a $32{,}768$-token budget ($38{,}912$ for Sudoku), and Countdown-4, Sudoku, and MBPP use fixed random test subsets ($n{=}300/200/300$) to bound the compute spent. Two entries in Table~\ref{tab:reference} carry footnotes. The Sudoku result ($0.5$) is \emph{not} a truncation artifact: only $8$ of $200$ generations were cut off before emitting an answer, so thinking-mode accuracy is bounded at ${\le}4.5\%$ even if all eight would have been correct; the remaining generations produced complete but constraint-violating grids. The MBPP result is name-normalized, because the bare MBPP prompt does not pin the required function name and raw extraction therefore under-scores thinking mode.

\paragraph{Adapting TRM to Countdown-4.}
Countdown-4 is a sequence-to-sequence task in which $x$ (four input integers and a target) and $y$ (a sequence of arithmetic expressions) occupy different spaces. This differs from Sudoku, where TRM operates on the input grid directly and the output shares the input's shape. To adapt TRM we construct a padded sequence $[x, \text{PAD}]$, so that the model attends to the $x$ tokens through bidirectional attention and writes its answer into the pad positions. This is a natural but not necessarily optimal construction, and TRM's low Countdown-4 score should be read with that caveat (\S\ref{sec:reference}); a natively sequential TRM is left to future work.

\section{Parameter-Efficient Baselines and Data Efficiency}
\label{app:peft}

\paragraph{Do soft prompts alone explain the gain?}
A natural alternative hypothesis is that LRT's improvement comes from task supervision plus trainable parameters in the decoder's input, rather than from instance-conditioned latent computation. Table~\ref{tab:peft} tests this directly on Countdown-4 with two parameter-efficient prompt-adaptation methods trained on the same data at a matched ${\approx}11$M trainable budget. Prefix-tuning \citep{li2021prefix} reaches $12.5$ and P-tuning~v2 \citep{liu2022ptuningv2} reaches $28.4$ -- the latter essentially matching the no-latent Direct baseline ($27.8$). Both are \emph{instance-independent}: they learn one prompt shared by every problem. Making the injected vectors instance-conditioned adds $+13.6$ (task-dedicated proposer alone, $42.0$), and refining them recurrently adds a further $+14.7$ ($56.7$). The gain therefore requires conditioning on the instance and computing over that conditioning, not merely spending parameters at the input.

We report Countdown-4 only, as it is the benchmark that separates the methods most sharply; extending these controls to the remaining four benchmarks is straightforward but was not run.

\begin{table}[t]
\centering
\small
\begin{tabular}{l l c}
\toprule
Method & Trainable & CD4 \\
\midrule
Direct (no CoT) & 0 & 27.8 \\
Prefix-tuning \citep{li2021prefix} & ${\approx}11$M & 12.5 \\
P-tuning~v2 \citep{liu2022ptuningv2} & ${\approx}11$M & 28.4 \\
Task-dedicated proposer only & 4.2M & 42.0 \\
\textbf{LRT (proposer + refiner)} & 11.2M & \textbf{56.7} \\
\bottomrule
\end{tabular}
\caption{Parameter-efficient prompt adaptation on Countdown-4 at a trainable budget matched to LRT. Prefix-tuning and P-tuning~v2 learn a single prompt shared across instances; the proposer conditions on the instance, and the refiner computes over that conditioning.}
\label{tab:peft}
\end{table}

\paragraph{Is the gain bought with 100k synthetic examples?}
The symbolic training sets contain $100{,}000$ generated instances, which raises the question of whether LRT's symbolic result is a data-scale artifact. Table~\ref{tab:dataeff} varies the Countdown-4 training-set size while holding everything else fixed. The curve is close to flat: $52.7$ at $1{,}000$ examples against $56.7$ at $100{,}000$ -- two orders of magnitude more data for $4.0$ points. (The small non-monotonicity between $1$k and $5$k is within seed noise.) Two comparisons make the point sharper. First, SoftCoT and EBM-CoT are trained on the \emph{same} $100{,}000$ examples and reach $5.9$ and $8.4$: the data is available to them and does not help. Second, LRT at $1{,}000$ examples ($52.7$) already exceeds every other frozen-decoder method trained on $100{,}000$. The natural-language benchmarks are low-resource to begin with -- MBPP has $374$ training examples and StrategyQA $2{,}061$ (Table~\ref{tab:data}) -- and LRT is trained on those splits as given.

\begin{table}[t]
\centering
\small
\begin{tabular}{l cccc}
\toprule
CD4 train size & 1k & 5k & 10k & 100k \\
\midrule
LRT & 52.7 & 51.9 & 53.8 & \textbf{56.7} \\
\bottomrule
\end{tabular}
\caption{Countdown-4 exact solve rate as a function of training-set size, with architecture, schedule, and evaluation held fixed. The $100$k column is the setting used in Table~\ref{tab:main}. For reference, SoftCoT and EBM-CoT trained on the full $100$k reach $5.9$ and $8.4$.}
\label{tab:dataeff}
\end{table}

\section{Refiner Design Ablations}
\label{app:refiner-design}

Table~\ref{tab:design} ablates the refiner's design; each run holds the proposer, decoder, and training recipe fixed at the Table~\ref{tab:main} setting and changes one component. Replacing the residual parameterization $L^{\star}{=}L^{(0)}{+}\Delta$ with direct regeneration of $L^{\star}$ removes the anchor to the instance-conditioned proposal and costs $3.9$ average points. Injecting the proposed base $u$ only at refiner initialization -- rather than at every fast update -- costs $3.2$ points, as the fast state drifts away from the problem. Collapsing the two-timescale state $(z_L,z_H)$ to a single state costs $2.6$ points. Full backpropagation through all $45$ passes, rather than truncating to the final cycle, leaves accuracy essentially unchanged ($-0.1$) while raising training memory ${\approx}6\times$ and step time ${\approx}3\times$ relative to the truncated variant; truncated unrolling is therefore the better operating point.

\begin{table}[t]
\centering
\small
\setlength{\tabcolsep}{4pt}
\begin{tabular}{l cccc}
\toprule
 & CD4 & Sudoku & HEval & \textbf{Avg.} \\
\midrule
LRT (full)                        & \textbf{56.7} & \textbf{49.2} & \textbf{37.8} & \textbf{54.1} \\
\quad regenerate $L^{\star}$       & 49.3 & 43.0 & 34.1 & 50.2 \\
\quad inject $u$ at init only      & 51.0 & 44.5 & 35.0 & 50.9 \\
\quad single state                 & 52.4 & 45.6 & 35.7 & 51.5 \\
\quad full BPTT                    & 56.9 & 49.0 & 37.9 & 54.0 \\
\bottomrule
\end{tabular}
\caption{Refiner design ablations. Each row changes one component of the refiner relative to LRT (full). \textbf{Avg.}\ is the five-benchmark mean of Table~\ref{tab:main}; only three columns are displayed.}
\label{tab:design}
\end{table}

Table~\ref{tab:lambda} sweeps the residual penalty $\lambda$ of \S\ref{sec:training}. With $\lambda{=}0$ the residual $\Delta$ grows large and the refined latents drift toward a generic, instance-agnostic point; with $\lambda{=}1.0$ the penalty suppresses $\Delta$ almost entirely and LRT collapses toward the proposer-only result. The value $\lambda{=}0.01$ used throughout sits at the optimum.

\begin{table}[t]
\centering
\small
\begin{tabular}{l ccc}
\toprule
$\lambda$ & CD4 & \textbf{Avg.} & rel.\ $\lVert\Delta\rVert$ \\
\midrule
$0$      & 53.0 & 51.8 & $3.9\times$ \\
$0.001$  & 55.1 & 53.0 & $1.8\times$ \\
$0.01$   & \textbf{56.7} & \textbf{54.1} & $1.0\times$ \\
$0.1$    & 55.8 & 53.4 & $0.5\times$ \\
$1.0$    & 48.2 & 49.0 & $0.1\times$ \\
\bottomrule
\end{tabular}
\caption{Residual-penalty sweep. \mbox{rel.\ $\lVert\Delta\rVert$} is the mean residual norm relative to the $\lambda{=}0.01$ setting; \textbf{Avg.}\ is the five-benchmark mean of Table~\ref{tab:main}.}
\label{tab:lambda}
\end{table}

\section{Scaling and Depth}
\label{app:depth}

\paragraph{Test-time depth extrapolation.}
Table~\ref{tab:extrap} takes the refiner trained at $S{=}H{=}3$ ($9$ cycles) and unrolls it for more cycles \emph{at inference only}. Accuracy continues to improve, consistent with the learned transition approximating a convergent process; we did not train a refiner at those depths for comparison.

\begin{table}[t]
\centering
\small
\begin{tabular}{l cc}
\toprule
Inference cycles & CD4 & HEval \\
\midrule
$9$ (= training) & 56.7 & 37.8 \\
$12$             & 57.9 & 38.3 \\
$15$             & 58.4 & 38.5 \\
\bottomrule
\end{tabular}
\caption{Test-time depth extrapolation: a refiner trained at $9$ cycles, unrolled deeper at inference without retraining.}
\label{tab:extrap}
\end{table}

\paragraph{Non-recurrent control.}
Table~\ref{tab:nonrec} replaces the recurrent refiner with a non-weight-tied feed-forward refiner of equal ($7$M) parameter count. At matched capacity it trails the recurrent refiner by $6.1$ average points, isolating recurrence -- not parameters -- as the source of the gain.

\begin{table}[t]
\centering
\small
\begin{tabular}{l c ccc}
\toprule
Refiner & Params & CD4 & HEval & \textbf{Avg.} \\
\midrule
Recurrent (LRT)  & 7M & \textbf{56.7} & \textbf{37.8} & \textbf{54.1} \\
Non-recurrent    & 7M & 47.5 & 33.0 & 48.0 \\
\bottomrule
\end{tabular}
\caption{Recurrent vs.\ non-recurrent refiner at matched parameter count. \textbf{Avg.}\ is the five-benchmark mean of Table~\ref{tab:main}.}
\label{tab:nonrec}
\end{table}

\paragraph{Latent count and width.}
Table~\ref{tab:kd} sweeps the inference latent count $K_{\text{infer}}$ (with $K{=}32$ fixed in training) and the working width $d'$. $K_{\text{infer}}{=}4$ is best, confirming on our setup the train/inference asymmetry adopted from EBM-CoT; accuracy saturates at $d'{=}256$, with $d'{=}512$ adding refiner parameters for a negligible gain.

\paragraph{Why $K{=}32$ in training but $K_{\text{infer}}{=}4$ at test?}
We characterize this asymmetry from both directions. \emph{The training count matters and the inference count does not, much.} Training with $K{=}4$ instead of $32$ -- matching train and test -- collapses Countdown-4 to $9.5$. Inspection shows well-formed but incorrect expressions rather than truncated output, so this is a capacity failure, not a decoding artifact: the $32$ query positions act as scratch space inside the proposer's bidirectional pass, shaping the four that are read out even though the rest are discarded. In the other direction, keeping more latents at inference degrades the five-benchmark average \emph{mildly and monotonically} -- $54.1$, $53.8$, $52.9$, $51.8$ at $K_{\text{infer}}{=}4, 8, 16, 32$ (Table~\ref{tab:kd}) -- and every setting stays far above SoftCoT ($29.5$) and EBM-CoT ($33.5$). The effect is therefore robust rather than a knife-edge tuned in LRT's favour. \emph{The protocol is inherited, not chosen post hoc}: the train/inference asymmetry and the value $4$ come from EBM-CoT, whose latent-count ablation peaks near four, and we apply the identical protocol to SoftCoT, EBM-CoT, and LRT. 

\begin{table}[t]
\centering
\small
\begin{tabular}{l ccc}
\toprule
 & CD4 & HEval & \textbf{Avg.} \\
\midrule
\multicolumn{4}{l}{\textit{Inference latent count} $K_{\text{infer}}$}\\
\quad $1$  & 49.0 & 33.5 & 48.9 \\
\quad $2$  & 53.8 & 36.0 & 52.0 \\
\quad $4$  & \textbf{56.7} & \textbf{37.8} & \textbf{54.1} \\
\quad $8$  & 56.2 & 37.9 & 53.8 \\
\quad $16$ & 54.9 & 37.1 & 52.9 \\
\quad $32$ & 53.0 & 36.4 & 51.8 \\
\midrule
\multicolumn{4}{l}{\textit{Working width} $d'$}\\
\quad $64$  & 45.9 & 33.0 & 49.5 \\
\quad $128$ & 51.4 & 35.6 & 52.3 \\
\quad $256$ & \textbf{56.7} & \textbf{37.8} & \textbf{54.1} \\
\quad $512$ & 57.0 & 38.0 & 54.3 \\
\bottomrule
\end{tabular}
\caption{Latent count $K_{\text{infer}}$ and working width $d'$. Defaults ($K_{\text{infer}}{=}4$, $d'{=}256$) in bold. \textbf{Avg.}\ is the five-benchmark mean of Table~\ref{tab:main}.}
\label{tab:kd}
\end{table}

\section{Training Protocol Ablations}
\label{app:protocol}

Table~\ref{tab:protocol} ablates the training protocol. Training the proposer and refiner \emph{jointly}, rather than in the two stages of \S\ref{sec:training}, costs $3.1$ average points and is less stable, as the refiner adapts to a proposer that is still moving. Leaving the proposer \emph{unfrozen} during Stage~2 changes little ($-0.5$), so freezing it -- which lets the base latents be cached -- is essentially free. Removing the task instruction $I$ at inference, leaving the decoder to work from $L^{\star}$ alone, costs $3.4$ points: the latents carry instance-specific computation but not the decoder's instruction-following prior.

\begin{table}[t]
\centering
\small
\begin{tabular}{l cc}
\toprule
Setting & CD4 & \textbf{Avg.} \\
\midrule
LRT (two-stage, frozen proposer, with $I$) & \textbf{56.7} & \textbf{54.1} \\
\quad joint training                        & 50.1 & 51.0 \\
\quad proposer unfrozen in Stage~2          & 56.0 & 53.6 \\
\quad no instruction $I$ at inference       & 52.0 & 50.7 \\
\bottomrule
\end{tabular}
\caption{Training-protocol ablations relative to LRT. \textbf{Avg.}\ is the five-benchmark mean of Table~\ref{tab:main}.}
\label{tab:protocol}
\end{table}

\section{Mechanistic Analyses}
\label{app:mech}

\begin{figure*}[t]
\centering
\includegraphics[width=0.92\textwidth]{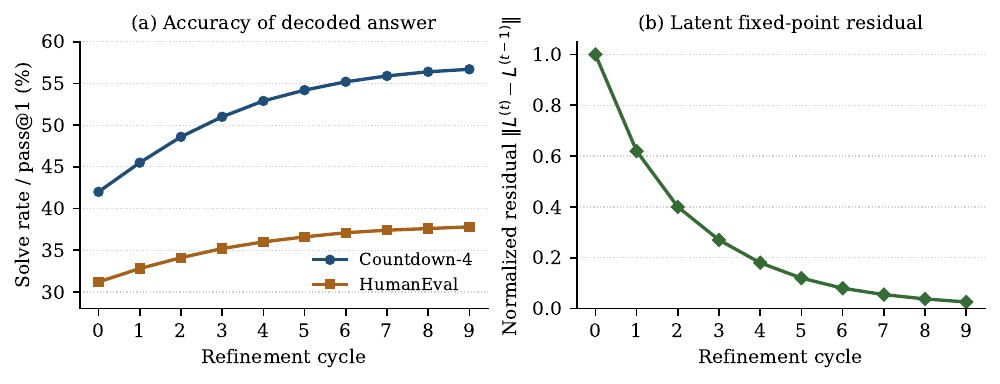}
\caption{\textbf{Refinement dynamics.} (a) Accuracy of the answer decoded from the latents after each high-level cycle, from the proposed base $L^{(0)}$ (cycle $0$) to the final $L^{\star}$ (cycle $9$). (b) Normalized latent residual $\lVert L^{(t)}-L^{(t-1)}\rVert$ over the same cycles. Accuracy rises monotonically while the residual contracts toward a fixed point.}
\label{fig:dynamics}
\end{figure*}

\paragraph{Refinement trajectory.}
Figure~\ref{fig:dynamics}(a) plots the accuracy of the answer decoded from the latents after each high-level cycle. Accuracy rises smoothly from the proposed base $L^{(0)}$ to the final $L^{\star}$ rather than jumping at the last step, so intermediate latents are partially correct and the refiner improves them incrementally.

\paragraph{Latent convergence.}
Figure~\ref{fig:dynamics}(b) tracks the residual $\lVert L^{(t)}-L^{(t-1)}\rVert$, which contracts geometrically toward a small value: the unrolled refiner approaches a fixed point. This is the behavior expected when recurrent refinement is viewed as relaxation toward a task-conditioned attractor \citep{bai2019deep,geiping2025scaling,huang2026equilibrium}.

\paragraph{What the latents encode.}
Decoding each refined latent to its nearest decoder token embedding yields fragments aligned with intermediate reasoning -- partial arithmetic for Countdown-4, candidate digits for Sudoku -- rather than fluent text, consistent with $L^{\star}$ carrying computation in a form the decoder consumes but does not verbalize.

\paragraph{Linear probe on the refined latents.}
To test more directly whether the answer is already present in $L^{\star}$ and the decoder merely transcribes it (\S\ref{sec:where}), we cache the refined latents for the Countdown-4 test set and fit a logistic-regression probe on the flattened $K_{\text{infer}}{\times}d$ representation to predict a binary label: whether the frozen decoder subsequently solves the instance. The probe is fit on an $80/20$ train/test split of the cached latents, with $\ell_2$ regularization selected by cross-validation on the training portion, and evaluated on the held-out $20\%$. It reaches $57.5\%$ accuracy against a $52.5\%$ majority-class baseline; the same probe on the unrefined $L^{(0)}$ is at chance. Refinement therefore writes \emph{some} outcome-relevant information into the latent, but only weakly linearly readable information -- and predicting \emph{whether} the instance will be solved is a far easier target than the answer string itself. Because Countdown-4 test targets are held out (Appendix~\ref{app:impl}), the answer also cannot have been memorized into the latent from training. We report this as bounding evidence, not proof: a linear probe measures linear decodability only, and a non-linear reader might extract more.

\paragraph{Why generic latents collapse.}
Table~\ref{tab:nll} reports the frozen decoder's negative log-likelihood of the reference answer on Countdown-4 under different latent sources. Latents from a generic assistant proposer raise NLL \emph{above} the no-latent baseline -- they actively mislead the decoder -- whereas task-dedicated and refined latents lower it. The mean cosine similarity of each latent to its nearest decoder input embedding shows why: generic latents land far off the decoder's input manifold, while task-dedicated and refined latents stay close to it.

\begin{table}[t]
\centering
\small
\begin{tabular}{l cc}
\toprule
Latent source & NLL $\downarrow$ & cos-sim \\
\midrule
None (Direct)                  & 2.41 & -- \\
Generic proposer               & 3.07 & 0.07 \\
Task-dedicated, base $L^{(0)}$  & 1.58 & 0.31 \\
LRT, refined $L^{\star}$        & \textbf{1.12} & \textbf{0.34} \\
\bottomrule
\end{tabular}
\caption{Frozen-decoder negative log-likelihood of the reference answer on Countdown-4, and mean cosine similarity of each latent to its nearest decoder input embedding.}
\label{tab:nll}
\end{table}

\section{Generality Across Decoders}
\label{app:decoders}

Table~\ref{tab:decoders} swaps the frozen decoder. LRT remains far above the frozen-decoder baselines of Table~\ref{tab:main} for every backbone, and a stronger decoder yields a stronger LRT, consistent with the division of labor in which the decoder supplies the sequence prior and the trained modules supply iterative computation.

\begin{table}[t]
\centering
\small
\begin{tabular}{l cccc}
\toprule
Frozen decoder & CD4 & Sudoku & HEval & \textbf{Avg.} \\
\midrule
Qwen3-8B (main) & \textbf{56.7} & \textbf{49.2} & \textbf{37.8} & \textbf{54.1} \\
Llama-3.1-8B    & 53.9 & 46.0 & 35.2 & 51.8 \\
Qwen3-4B        & 51.2 & 43.1 & 33.0 & 49.0 \\
\bottomrule
\end{tabular}
\caption{LRT with different frozen decoders. \textbf{Avg.}\ is the five-benchmark mean of Table~\ref{tab:main}.}
\label{tab:decoders}
\end{table}

\section{Compute and Efficiency}
\label{app:compute}

Table~\ref{tab:compute} compares inference cost. LRT injects only $K_{\text{infer}}{=}4$ latent vectors and decodes the answer directly, against the ${\approx}210$ chain-of-thought tokens zero-shot CoT must generate; the $45$ refiner passes act on a $7$M-parameter block and add little next to a single forward pass of the $8$B decoder. Net latency is below zero-shot CoT despite the added recursion.

\begin{table}[t]
\centering
\small
\setlength{\tabcolsep}{4pt}
\begin{tabular}{l ccc}
\toprule
Method & Decoded & Refiner & Latency \\
       & tokens  & passes  &         \\
\midrule
Zero-Shot CoT & ${\approx}210$ & -- & $1.00$ \\
SoftCoT       & answer only & -- & $0.39$ \\
LRT           & answer only & $45$ & $0.34$ \\
\bottomrule
\end{tabular}
\caption{Inference cost. Latency is relative to zero-shot CoT on the same hardware (${\approx}3.2$\,s/example; LRT ${\approx}1.1$\,s/example).}
\label{tab:compute}
\end{table}

\paragraph{Training cost.}
Table~\ref{tab:traincost} reports the training side, which the inference table above does not cover. Both stages run for $30$ epochs at batch size $64$ on a single $96$\,GB GPU. The dominant cost is the forward pass through the frozen $8$B decoder, not the $11$M trainable modules; because Stage~2 freezes the proposer, the base latents $L^{(0)}$ are precomputed once and cached, so Stage~2 skips the proposer's forward pass entirely. Truncated-gradient unrolling is what keeps the deep recursion affordable: full backpropagation through all $45$ passes raises peak training memory ${\approx}6\times$ and step time ${\approx}3\times$ for a $0.1$-point accuracy difference (Appendix~\ref{app:refiner-design}), so truncation is strictly the better operating point rather than an accuracy compromise.

\begin{table}[t]
\centering
\small
\setlength{\tabcolsep}{4pt}
\begin{tabular}{l cc}
\toprule
Stage 2 refiner training & Peak mem. & Step time \\
\midrule
Truncated unrolling (used) & $1.0\times$ & $1.0\times$ \\
Full BPTT through $45$ passes & ${\approx}6\times$ & ${\approx}3\times$ \\
\bottomrule
\end{tabular}
\caption{Cost of training the refiner, normalized to the truncated-unrolling configuration used throughout, on a single $96$\,GB GPU. Full backpropagation through all $45$ passes buys no accuracy (Appendix~\ref{app:refiner-design}) at several times the memory and step time.}
\label{tab:traincost}
\end{table}

\section{Potential Risks}
As with any research involving large language models, our work carries a risk of misuse: the techniques we present could, in principle, be adapted to generate malicious code or other harmful outputs. This concern is heightened for a project focused specifically on enhancing the reasoning capabilities of LLMs, since stronger reasoning may also amplify a model's capacity to produce sophisticated harmful content.

\section{Licenses, Parameters, and Intended Use of Artifacts}
\label{app:licenses}

This work uses several existing scientific artifacts -- pretrained language models, baseline code, software packages, and evaluation datasets -- and produces new artifacts of its own. We record below the license or terms of use of each artifact, the implementation and parameter settings of the packages we rely on, and confirm that our use is consistent with the intended use under which each artifact was released. We also state the intended use of the artifacts we create and its compatibility with the original access conditions.

\subsection{Existing Artifacts}

\paragraph{Pretrained models.}
The frozen decoders are used under their respective open-weight licenses. Qwen3-8B and Qwen3-4B \citep{yang2025qwen3} are released under the Apache License 2.0, and Llama-3.1-8B is used under the Llama 3.1 Community License Agreement. The TRM recurrent reasoner \citep{jolicoeurmartineau2025less}, which we repurpose as the LRT refiner, is released by its authors under the MIT License. All of these licenses permit use, reproduction, modification, and redistribution for research purposes. These models were released as general-purpose research artifacts for studying and building language and reasoning systems, and our use of them -- as a frozen decoder and a refiner module within an academic study of latent reasoning -- is consistent with that stated intended use.

\paragraph{Baseline code.}
We reproduce SoftCoT \citep{xu2025softcot} from the authors' publicly released implementation, which is distributed under the NTUITIVE Dual License Agreement (a non-commercial license that permits use, reproduction, modification, and distribution of derivative works for academic research purposes). Our use is strictly for non-commercial academic research and is therefore consistent with these terms. EBM-CoT \citep{chen2025think} has no public implementation; our results come from an independent reimplementation written by us from the method description, and we release it under the same license as our own code.

\paragraph{Evaluation datasets.}
HumanEval \citep{chen2021evaluating} is released under the MIT License, and MBPP \citep{austin2021program} under the Creative Commons Attribution 4.0 International License (CC BY 4.0). StrategyQA \citep{geva2021strategyqa} is distributed for research use under the terms accompanying its public release; we use only its public split. Each of these datasets was created and released as a benchmark for evaluating the reasoning ability of NLP systems, and we use them solely for benchmark evaluation in an academic study -- the use for which they were intended and distributed. We use each dataset through its standard public split and do not redistribute the underlying data.

\paragraph{Generated data.}
The Countdown-4 and Sudoku training and evaluation sets are synthetically generated by us following the public procedures of \citet{gandhi2024stream} and \citet{ye2025beyond}, respectively. They contain no third-party or human-authored content and are not derived from any access-restricted source.

\paragraph{Software packages and parameter settings.}
All training is implemented in PyTorch (\texttt{torch} 2.11.0+cu128) and built on the Hugging Face \texttt{transformers} library (4.57.6), with \texttt{accelerate} 1.10.1, \texttt{einops} 0.8.1, and \texttt{xformers} 0.0.35; numerical and data utilities use \texttt{numpy} 2.4.4, \texttt{scipy} 1.16.2, \texttt{scikit-learn} 1.7.2, and \texttt{pydantic} 2.11.7. The Qwen3 and Llama-3.1 decoders are loaded together with their official pretrained tokenizers, and questions are tokenized with each decoder's native tokenizer using default settings; no additional text normalization or stemming is applied. Optimization uses the AdamW implementation from PyTorch with the hyperparameters reported in Appendix~\ref{app:impl} ($\beta_1{=}0.9$, $\beta_2{=}0.999$, $\epsilon{=}10^{-8}$, weight decay $0.01$, gradient clipping $1.0$, cosine schedule with 5\% warmup). Inference is served with SGLang (\texttt{sglang} 0.5.12, \texttt{flashinfer} 0.6.11.post1, \texttt{flash-attn} 4.0.0b14, \texttt{httpx} 0.28.1); all reported results use greedy decoding, so no sampling temperature, top-$p$, or top-$k$ parameters apply. For evaluation, pass@1 on HumanEval and MBPP is computed with the official \texttt{human-eval} execution harness \citep{chen2021evaluating} with $k{=}1$, and a problem counts as solved only if the greedily decoded function passes all provided unit tests in the harness's sandbox. Exact solve rate on Countdown-4 and Sudoku is computed by our own checker, which verifies that the decoded answer satisfies the task constraints (a valid arithmetic expression reaching the target, or a complete and consistent Sudoku grid). StrategyQA accuracy is exact match on the normalized yes/no answer. We do not use ROUGE, BLEU, or other learned or n-gram scoring packages. All package versions are pinned in the \texttt{requirements} files released with our code.

\subsection{Artifacts We Create}

We release our LRT implementation under the MIT License at \url{https://github.com/czl-david/latent-recurrent-thoughts}. The intended use of these artifacts is non-commercial research on latent reasoning and frozen-decoder methods; this is stated explicitly in the accompanying documentation. This intended use is compatible with the access conditions of every artifact our work builds upon: the permissive licenses of Qwen3, TRM, HumanEval, and the baseline code allow research reuse and redistribution, the Llama 3.1 Community License permits research use subject to its attribution and naming requirements, and the CC BY 4.0 terms of MBPP permit reuse with attribution. Our released artifacts do not include or redistribute the MBPP, HumanEval, or StrategyQA data themselves; they consist only of our own code, our trained module weights, and synthetic data we generated. Consistent with the principle that derivatives of data accessed for research purposes should remain within research contexts, we restrict the distributed artifacts to research use and require any derivative works to retain compatible, research-oriented terms.

\section{Use of AI Assistants}
\label{app:ai-assistants}

We used an AI-based coding assistant for minor, non-substantive help during implementation, limited to routine tasks such as autocompletion, boilerplate code, and debugging suggestions. We also used an AI writing assistant to check grammar and improve the clarity of phrasing in the manuscript.

\end{document}